\PassOptionsToPackage{table}{xcolor}
\documentclass[11pt]{article}

\usepackage[preprint]{acl}

\usepackage[utf8]{inputenc} 
\usepackage[T1]{fontenc}    
\usepackage{times}          
\usepackage{latexsym}       
\usepackage{booktabs}       
\usepackage{amsfonts}       
\usepackage{amsmath}        
\usepackage{nicefrac}       
\usepackage{microtype}      
\usepackage{xcolor}         
\usepackage{booktabs}
\usepackage{multirow}
\usepackage{graphicx}
\usepackage{pifont}
\usepackage{colortbl}       
\usepackage{enumitem}       

\newcommand{\benchmark}{EntSQL}

\title{EntSQL: A Benchmark for Grounding Text-to-SQL in Long-Context Enterprise Knowledge}

\author{%
\begin{tabular}{@{}c@{}}
Chengxi Liao\textsuperscript{1,*} \quad
Tao Xu\textsuperscript{2,*} \quad
Zulong Chen\textsuperscript{2,\ensuremath{\dagger}} \quad
Chuanfei Xu\textsuperscript{4} \quad
Yiyan Wang\textsuperscript{2} \\
Xinyun Wang\textsuperscript{2} \quad
Yanlong Zhang\textsuperscript{2} \quad
Xiaojun Chen\textsuperscript{2} \quad
Zhibo Yang\textsuperscript{2,3} \quad
Zeyi Wen\textsuperscript{1,\ensuremath{\dagger}} \\
\\[-0.3em]
{\normalfont\normalsize
\textsuperscript{1}HKUST (GZ) \quad
\textsuperscript{2}Alibaba Group \quad
\textsuperscript{3}Qwen Team} \\
{\normalfont\normalsize
\textsuperscript{4}Guangdong Laboratory of Artificial Intelligence and Digital Economy (SZ)} \\
{\normalfont\footnotesize
\href{mailto:zulong.czl@alibaba-inc.com}{%
  {\color{black}\texttt{zulong.czl@alibaba-inc.com}}%
},
\href{mailto:wenzeyi@hkust-gz.edu.cn}{%
  {\color{black}\texttt{wenzeyi@hkust-gz.edu.cn}}%
}}
\end{tabular}
}

\begin{document}

\maketitle
\begingroup
\makeatletter
\renewcommand{\@makefnmark}{}
\footnotetext[1]{
\textsuperscript{*}Equal contribution.
}
\footnotetext[1]{
\textsuperscript{\ensuremath{\dagger}}Corresponding authors.
}
\makeatother
\endgroup

\begin{abstract}
Text-to-SQL enables natural language access to databases, and recent LLMs have substantially advanced its capabilities. 
Existing benchmarks such as Spider, BIRD, and Spider~2.0 evaluate schema generalization, large-scale databases, and realistic workflows, but largely overlook enterprise scenarios where SQL generation depends on private business knowledge, such as internal metrics, reporting conventions, and organizational rules. We introduce \benchmark{}, an enterprise-oriented Text-to-SQL benchmark for evaluating long-context grounding over proprietary business documents. \benchmark{} contains 1,066 aligned Chinese-English semantic examples across five business domains, with most examples requiring domain knowledge beyond the question and schema and involving complex SQL structures. On English inputs, the best evaluated system reaches only 15.9\% when long-form documents are provided, highlighting the difficulty of grounding SQL generation in enterprise knowledge.
The benchmark is publicly available at \url{https://huggingface.co/datasets/XuWave/Rethinking_Enterprise_Text_to_SQL}.
\end{abstract}

\section{Introduction}

Text-to-SQL, the task of translating natural language questions into executable SQL queries, has emerged as a pivotal technology for democratizing data access~\citep{liu2025survey}. By enabling non-technical users to query databases without manual coding, Text-to-SQL systems bridge the gap between natural human intent and structured relational data, making data-driven business insights accessible to a broader audience beyond professional data teams. The advent of Large Language Models (LLMs) has revolutionized this field, transforming Text-to-SQL from a narrow specialized semantic parsing task into a comprehensive multi-step reasoning challenge that encompasses schema understanding, cross-domain semantic alignment, and complex nested query decomposition~\citep{pourreza2023din, gao2023text, dong2023c3}. In enterprise business intelligence (BI) settings, these technical advances have substantially improved the real-world deployment feasibility of Text-to-SQL systems for end-to-end business analytics workflows.

Existing Text-to-SQL benchmarks have progressively broadened the evaluation landscape. Spider introduced a cross-domain semantic parsing benchmark that stresses schema generalization across unseen databases~\citep{yu2018spider}. BIRD moves evaluation further toward realistic database environments by incorporating large-scale databases, external knowledge, and efficiency-oriented evaluation~\citep{li2023can}. Spider~2.0 extends this trajectory to complex real-world workflows with larger schemas, external tools, and public documentation in industrial-style settings~\citep{lei2024spider}. Together, these benchmarks cover important aspects of schema generalization, database grounding, short-form evidence use, and industrial-scale workflow execution. However, realistic enterprise question-answering workflows also involve private enterprise-oriented content, such as internal metric definitions, reporting conventions, fiscal-year rules, organization and product mappings, and management-specific adjustment policies. Consider a representative enterprise query:
\begin{quote}
    \itshape ``Compare FY2024 and FY2025 revenue growth for top 10 products and analyze how each cost item changed.''
\end{quote}
Resolving such a query may require interpreting internal reporting conventions, metric definitions, temporal rules, or entity mappings that are not recoverable from public knowledge alone. The challenge is therefore not merely to write syntactically valid SQL, but to ground SQL construction in long-context enterprise business knowledge.

To study this setting, we introduce \benchmark{}, a new enterprise-oriented Text-to-SQL benchmark designed to evaluate whether models can ground SQL generation in long-form private business knowledge. \benchmark{} contains 1,066 aligned bilingual semantic examples across five enterprise BI domains: finance, treasury, business management, party building, and human resources. Each instance pairs a natural language question and database with a long-form enterprise document that describes the relevant business context. 
The benchmark is designed to simulate realistic enterprise question-answering scenarios and reflect private business rules rather than open-domain facts, following established industrial deployment practices for enterprise BI systems: 96.0\% of examples require domain knowledge beyond the question and schema, and the gold SQL queries average 388.7 tokens with frequent Common Table Expressions (CTEs), aligning with the complexity of real-world enterprise SQL workloads.

In summary, our main contribution is \benchmark{}, which complements existing Text-to-SQL benchmarks by focusing on enterprise long-context grounding with aligned Chinese and English inputs. 
Unlike benchmarks that primarily emphasize schema generalization, database grounding, or public-documentation workflows~\citep{yu2018spider, li2023can, lei2024spider}, \benchmark{} evaluates whether models can use previously unseen private enterprise knowledge to generate executable SQL. 
We evaluate strong standalone models and an interactive coding agent on the English version of \benchmark{}, using execution accuracy as the primary metric~\citep{yu2018spider}. 
The best-performing evaluated system reaches only 15.9\% when long-form documents are provided, indicating that enterprise long-context Text-to-SQL remains a substantial challenge for current state-of-the-art systems~\citep{liu2025survey}.

\section{Our Benchmark}
\label{sec:benchmark}

The goal of \benchmark{} is to evaluate Text-to-SQL systems in realistic enterprise question-answering workflows that are not specifically targeted by existing benchmarks. In such workflows, users ask business intelligence questions over private proprietary workloads, where SQL generation may depend on enterprise-oriented content such as finance, treasury, business management, party building, and human resources. To reflect this setting, we construct \benchmark{} from real enterprise operational data and organize the benchmark around private business rules, internal reporting conventions, and domain-specific analytical needs. In this section, we first define the task formulation and evaluation metric, then describe the data construction pipeline, and finally characterize the benchmark composition.

\subsection{Task Formulation}
\label{subsec:task-formulation}

\paragraph{Problem Definition.}
Given a natural language question $Q$, a database schema $S$, and a long-form domain document $D$, the task is to generate an executable SQL query $\hat{Y}$ that matches user intent. Let $\mathcal{R}_{\mathcal{DB}}(Y)$ denote the result set returned by executing SQL query $Y$ on database $\mathcal{DB}$. A prediction $\hat{Y}_i$ is correct if
\begin{equation}
\mathcal{R}_{\mathcal{DB}_i}(\hat{Y}_i) \equiv
\mathcal{R}_{\mathcal{DB}_i}(Y_i^{*}),
\end{equation}
where $Y_i^{*}$ denotes the gold SQL query and $\equiv$ denotes order-insensitive result-set equivalence.

\paragraph{Evaluation Metrics.}
We use \textbf{execution accuracy} (EX) as the primary metric, following recent Text-to-SQL benchmark practice that compares the denotation of a predicted SQL query against the gold query on the target database~\citep{li2023can, lei2024spider}. Compared with exact string matching, execution-based evaluation better captures semantic equivalence of different SQL surface forms. For a dataset of $N$ queries:
\begin{equation}
\mathrm{EX} = \frac{1}{N}\sum_{i=1}^{N}
\mathbf{1}\!\left[
\mathcal{R}_{\mathcal{DB}_i}(\hat{Y}_i) \equiv
\mathcal{R}_{\mathcal{DB}_i}(Y_i^{*})
\right].
\end{equation}

\subsection{Data Construction Pipeline}
\label{subsec:construction}

\begin{figure*}[t]
\centering
\includegraphics[width=\textwidth]{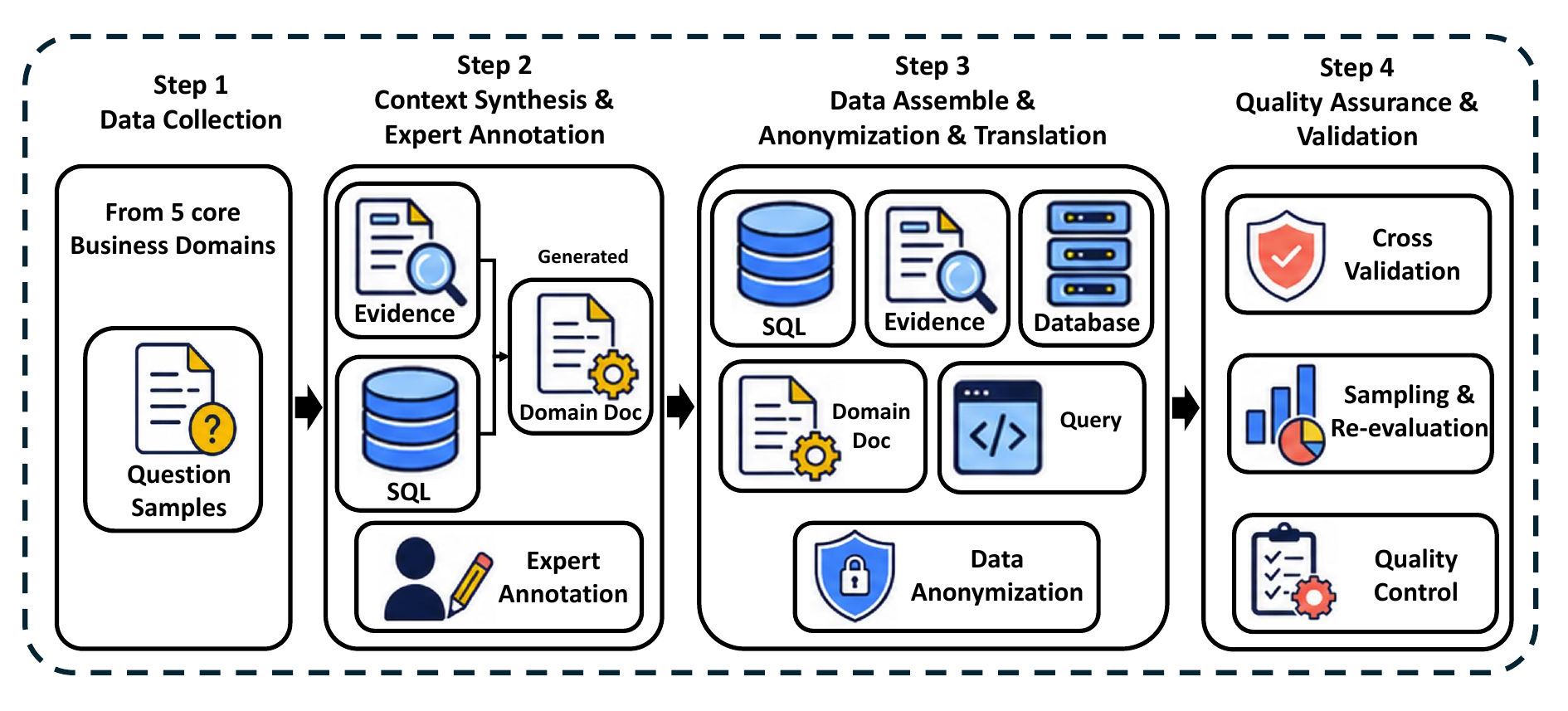}
\caption{Data construction pipeline. We collect enterprise queries, annotate SQL and domain evidence, anonymize sensitive content, and validate all annotations through multi-stage quality control.}
\label{fig:pipeline}
\end{figure*}

Our benchmark construction follows a central principle: each instance should require not only schema linking, but also the assimilation of enterprise-specific rules that are unavailable from public knowledge. Guided by this principle, we construct the benchmark through four stages, as shown in Figure~\ref{fig:pipeline}, covering authentic query collection, expert-grounded context synthesis, privacy-preserving anonymization, and multi-stage quality control.

\textbf{Data Collection.}
We collect real user Q\&A logs from multiple enterprise domains to ensure that the benchmark reflects genuine business information needs rather than artificially designed examples.
The data covers five major workflow categories, including treasury management, financial analytics, human resources, business management, and organizational operations.
We retain complex analytical queries involving multi-table joins, nested subqueries, aggregations, and temporal comparisons, which are common in enterprise BI scenarios and require non-trivial business reasoning.

\textbf{Context Synthesis and Expert Annotation.}
Enterprise queries often depend on domain-specific knowledge that is not explicit in the question or schema.
To capture such knowledge, domain experts annotate each instance with three components: the gold SQL reflecting the intended business logic, concise evidence snippets identifying the key business rules, and a synthesized domain document built from the evidence and supporting business materials.
Each instance therefore contains a question, SQL, evidence, domain document, and database, enabling evaluation of domain-aware and long-context Text-to-SQL reasoning.

\begin{table*}[t]
\centering

\small  
\setlength{\tabcolsep}{5pt} 

\resizebox{0.98\textwidth}{!}{
\begin{tabular}{l r c c r r r}
\toprule
\textbf{Dataset} & \textbf{\#data} & \textbf{Avg. Q} & \textbf{Avg. Doc} & \textbf{Easy} & \textbf{Medium} & \textbf{Hard} \\
\midrule
\textbf{Treasury} & \textbf{441} & \textbf{54.8} & \textbf{3.6k} & \textbf{5} & \textbf{255} & \textbf{181} \\
\quad Operating Capital \& Transactions & 110 & 55.2 & 3.6k & 0 & 55 & 55 \\
\quad Balance \& Account Financing & 86 & 58.3 & 3.5k & 1 & 43 & 42 \\
\quad Settlement \& Account Association & 72 & 49.2 & 3.6k & 3 & 53 & 16 \\
\quad Liquidity Management & 103 & 54.2 & 3.6k & 1 & 63 & 39 \\
\quad Accounts \& Capital Totals & 70 & 56.4 & 3.6k & 0 & 41 & 29 \\
\midrule
\textbf{Finance} & \textbf{351} & \textbf{39.8} & \textbf{4.4k} & \textbf{16} & \textbf{166} & \textbf{169} \\
\quad Profit & 87 & 33.2 & 4.7k & 8 & 19 & 60 \\
\quad Revenue & 137 & 39.2 & 4.0k & 3 & 62 & 72 \\
\quad EBITA & 127 & 45.0 & 4.7k & 5 & 85 & 37 \\
\midrule
\textbf{Human Resources} & \textbf{124} & \textbf{32.8} & \textbf{5.3k} & \textbf{109} & \textbf{15} & \textbf{0} \\
\quad Basic Personnel & 40 & 37.1 & 5.8k & 36 & 4 & 0 \\
\quad Onboarding/Offboarding & 20 & 22.8 & 6.1k & 20 & 0 & 0 \\
\quad Performance Review & 32 & 35.5 & 3.5k & 21 & 11 & 0 \\
\quad Promotion & 8 & 38.8 & 3.4k & 8 & 0 & 0 \\
\quad Benefits & 24 & 28.1 & 5.3k & 24 & 0 & 0 \\
\midrule
\textbf{Business Management} & \textbf{100} & \textbf{29.5} & \textbf{5.1k} & \textbf{18} & \textbf{73} & \textbf{9} \\
\quad Management & 100 & 29.5 & 5.1k & 18 & 73 & 9 \\
\midrule
\textbf{Union/Party Building} & \textbf{50} & \textbf{33.3} & \textbf{4.7k} & \textbf{0} & \textbf{33} & \textbf{17} \\
\quad Union & 50 & 33.3 & 4.7k & 0 & 33 & 17 \\
\bottomrule
\end{tabular}
}
\caption{Domain-wise data statistics and SQL difficulty. \#data denotes aligned bilingual examples; question and document lengths are average English token counts, with document averages computed over available documents.}
\label{tab:subdomain}
\end{table*}

\textbf{Data Assembly, Anonymization, and Translation.}
After annotation, we assemble the components into complete benchmark instances and apply multi-level anonymization to protect enterprise-sensitive information.
For questions and SQL queries, sensitive entities such as dates, monetary values, organizational units, and financial institutions are replaced using manually curated mappings while preserving query structure and reasoning patterns.
For databases, we anonymize dimension tables first and propagate the mappings to fact tables through join keys to maintain referential integrity.
Evidence and domain documents are also manually reviewed to prevent leakage of proprietary business logic or sensitive operational knowledge.
The original Chinese questions, evidence, and documents are further translated and validated into English, producing paired Chinese--English inputs aligned to the same database and gold SQL.

\textbf{Quality Assurance and Validation.}
To ensure dataset reliability, all instances undergo multiple rounds of manual review by domain experts, business specialists, and product practitioners.
Reviewers verify the consistency among questions, SQL annotations, evidence, domain documents, and database content.
We also conduct sampling-based independent validation to assess annotation quality and identify potential inconsistencies.
This process improves data fidelity and supports reliable evaluation in complex enterprise scenarios.

\begin{figure*}[t]
\vspace{-6pt}
\centering

\begin{minipage}{0.35\textwidth}
\centering
\includegraphics[width=\linewidth]{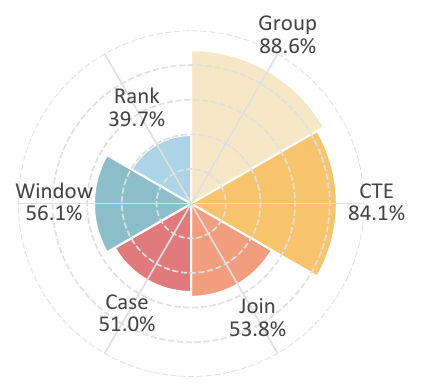}
\end{minipage}
\hfill
\begin{minipage}{0.31\textwidth}
\centering
\includegraphics[width=\linewidth]{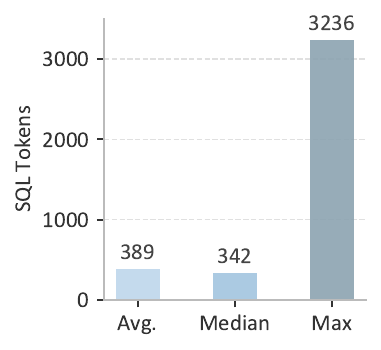}
\end{minipage}
\hfill
\begin{minipage}{0.31\textwidth}
\centering
\includegraphics[width=\linewidth]{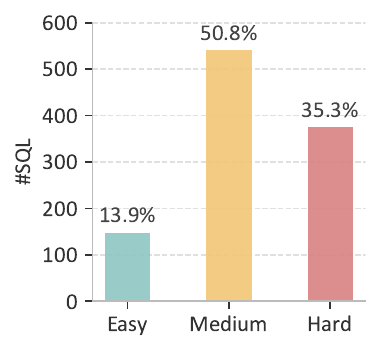}
\end{minipage}

\vspace{-6pt}
\caption{Gold SQL profile in \benchmark{}. From left to right: feature distribution, gold SQL token-length statistics, and token-length-based difficulty distribution.}
\label{fig:sql-profile-bars}
\vspace{-10pt}
\end{figure*}

\subsection{Benchmark Composition}
\label{subsec:statistics}

\benchmark{} includes 1,066 aligned bilingual semantic examples spanning five domains and 15 subdomains, with 15 database groups, 35 tables, and 1,489 columns from real-world enterprise workloads. The domain-wise question and document length statistics are computed over English token counts. Unless otherwise stated, token counts in this section are computed with OpenAI's \texttt{cl100k\_base} tokenizer. Notably, 96.0\% of examples require external domain knowledge beyond the question and schema.
Table~\ref{tab:subdomain} further breaks down domain-level statistics: Treasury and Finance dominate both scale and complexity, while Human Resources involves smaller schemas and shorter SQL but still relies on domain-specific terminology and reporting conventions. Figure~\ref{fig:sql-profile-bars} presents the overall gold SQL profile.

To characterize query complexity, we categorize examples into three difficulty levels based on gold SQL tokens: \textit{Easy} (<200 tokens), \textit{Medium} (200--400 tokens), and \textit{Hard} (>400 tokens). While coarse, SQL token length serves as a practical proxy for query complexity in our benchmark. The average SQL length is 388.7 tokens, with most queries falling into the Medium and Hard categories, reflecting multi-step reasoning in enterprise BI scenarios.

\section{Experiments}
\label{sec:experiments}

\subsection{Experimental Setup}
\label{subsec:setup}

\textbf{Benchmark Setting.}
We evaluate all systems on the full English \benchmark{} benchmark. Each example is associated with a target database, and most examples are associated with a domain-specific document. For Base, the model receives the question; for w/Doc, it receives the question and long-form document. At test time, models are provided with a natural language question, database schema, and optionally a long-form domain document, while gold SQL annotations are kept private. The task requires translating business questions into executable SQL queries, often relying on knowledge beyond the database schema.

\textbf{Models.} 
We evaluate eight systems, including seven standalone large language models and one interactive agent. The standalone models are Claude Opus 4.6 (Opus 4.6), Claude Sonnet 4.6 (Sonnet 4.6)~\citep{anthropic2026claude46}, GPT-5.4~\citep{openai2026gpt54}, Gemini 3.1 Pro Preview (Gemini 3.1 Pro)~\citep{google2026gemini31}, Qwen 3.6 Max Preview (Qwen 3.6 Max)~\citep{qwen2026qwen36max}, Kimi K 2.6 (Kimi K2.6)~\citep{moonshot2026kimik26}, and GLM 5.1~\citep{zeng2026glm}. They are evaluated in a single-turn setting through unified Python scripts. 

For the agent setting, we use Claude Code powered by Claude Sonnet 4.6 (Claude Code) and its default coding-agent framework~\citep{anthropic2025claudecode}. Although designed for coding, Claude Code shares the core workflow of modern NL2SQL agents, including tool use, execution-based validation, feedback-driven refinement, and multi-turn reasoning. We therefore use Claude Code as a practical proxy for NL2SQL agent workflows. All outputs are evaluated without manual correction.

\textbf{Input Settings.}
We consider three benchmark input settings. \textit{Question Only} includes the natural language question and schema, testing schema linking and parametric knowledge. \textit{Question+Doc} additionally provides the associated long-form domain document, reflecting the full benchmark setting. \textit{Question+Evidence} provides concise evidence snippets when available and is used only for ablation analysis of targeted knowledge grounding.

\textbf{Evaluation Protocol.}
All SQL queries are executed against local SQLite database files using the Python standard-library \texttt{sqlite3} backend. We report \texttt{pass@1} execution accuracy as the primary metric: each example contributes one final predicted SQL query, whose execution result is compared with the gold SQL result. Results are considered correct if the outputs are equivalent after normalization, such as ignoring ordering when not specified. For standalone model scripts, model calls use temperature 0, maximum output length is set to 8192 tokens, and API requests use a timeout of 300 seconds. Each example is evaluated through a single-turn model call, with up to three wrapper-level retries per example to handle failed calls or malformed responses; these infrastructure retries do not sample multiple candidate SQL programs for scoring. Claude Code is run separately with its default framework settings and temperature 0 where exposed; we do not manually set a max-turn budget or budget-token setting.

\textbf{Implementation Details.} 
Database schemas are serialized into textual form using DDL-style table and column information. A unified prompt template is applied to the standalone model scripts, with schema and (if applicable) document or evidence concatenated into the input. Claude Code receives the same task inputs but is run through its default framework configuration. wrapper-level retries are used only for standalone model API failures or malformed responses, not for Claude Code interactive self-correction.


\begin{table*}[t]
\centering
\footnotesize
\setlength{\tabcolsep}{2.8pt}
\renewcommand{\arraystretch}{1.12}
\resizebox{\textwidth}{!}{%
\begin{tabular}{l cccc @{\hspace{0.5em}}|@{\hspace{0.5em}} cccc @{\hspace{0.5em}}|@{\hspace{0.5em}} cccc}
\toprule
\multirow{2}{*}{\textbf{Systems}} &
\multicolumn{4}{c}{\textbf{Base}} &
\multicolumn{4}{c}{\textbf{w/Doc}} &
\multicolumn{4}{c}{\textbf{w/Evidence}} \\
\cmidrule(lr){2-5} \cmidrule(lr){6-9} \cmidrule(lr){10-13}
& \textbf{Easy} & \textbf{Med.} & \textbf{Hard} & \textbf{Avg}
& \textbf{Easy} & \textbf{Med.} & \textbf{Hard} & \textbf{Avg}
& \textbf{Easy} & \textbf{Med.} & \textbf{Hard} & \textbf{Avg} \\
\midrule
Claude Code 
& \textbf{27.7} & \textbf{5.5} & 0.3 & \textbf{6.8}
& \textbf{29.1} & \textbf{18.3} & \textbf{7.2} & \textbf{15.9}
& 33.8 & \textbf{25.6} & \textbf{10.4} & \textbf{21.4} \\

\rowcolor{gray!15}
Sonnet 4.6 
& 23.6 & 4.8 & 0.5 & 5.9
& 24.3 & 13.5 & 4.0 & 11.6
& 28.4 & 22.5 & 6.4 & 17.6 \\

Opus 4.6 
& \textbf{27.7} & 3.5 & \textbf{0.8} & 5.9
& \textbf{29.1} & 14.9 & 4.8 & 13.3
& 33.8 & 15.7 & 8.0 & 15.5 \\

\rowcolor{gray!15}
Gemini 3.1 Pro 
& 22.3 & 3.0 & 0.3 & 4.7
& 23.6 & 10.9 & 3.2 & 9.9
& 25.0 & 15.3 & 4.3 & 12.8 \\

GPT-5.4 
& 25.7 & 2.4 & 0.0 & 4.8
& 25.7 & 12.9 & 2.9 & 11.2
& 30.4 & 18.8 & 8.2 & 16.7 \\

\rowcolor{gray!15}
Qwen 3.6 Max 
& \textbf{27.7} & 3.7 & 0.0 & 5.7
& 27.0 & 13.3 & 5.9 & 12.6
& \textbf{35.1} & 15.5 & 10.1 & 16.3 \\

Kimi K2.6 
& 21.6 & 3.5 & 0.0 & 4.8
& 23.0 & 9.4 & 0.8 & 8.3
& 22.3 & 11.6 & 4.5 & 10.6 \\

\rowcolor{gray!15}
GLM 5.1 
& 27.0 & 3.5 & 0.0 & 5.5
& 27.7 & 15.1 & 4.0 & 12.9
& 29.7 & 18.8 & 9.6 & 17.1 \\
\bottomrule
\end{tabular}
}
\caption{
Execution accuracy (\%) on English examples by SQL-length-based difficulty group and input setting.
\textbf{Med.} denotes Medium.
\textbf{Avg} denotes weighted accuracy over all 1,066 English examples.
\textbf{Base} denotes Question Only, \textbf{w/Doc} adds long-form domain documents, and \textbf{w/Evidence} provides concise expert-curated evidence.
Best results within each column are boldfaced.
}
\label{tab:main}
\end{table*}


\subsection{Evaluation Results}
\label{subsec:evaluation-results}

Table~\ref{tab:main} reports execution accuracy on English inputs across SQL-length-based difficulty groups under the Base, w/Doc, and w/Evidence settings.

\textbf{(1) Performance drops sharply as SQL complexity increases.}
All systems show a large accuracy decline from Easy to Medium and Hard queries, indicating that current models struggle with longer and more compositional SQL programs.
In the Base setting, the best Hard accuracy is only 0.8\%, suggesting that questions and schemas alone are largely insufficient for complex enterprise SQL generation.

\textbf{(2) Concise evidence is more effective than full documents.}
Adding long-form domain documents improves the best Avg accuracy from 6.8\% to 15.9\%, while providing expert-curated evidence further increases it to 21.4\%.
This gap between w/Doc and w/Evidence suggests that enterprise knowledge is useful when it is localized, but current models still struggle to identify and apply relevant information from long documents.
Thus, effective knowledge selection and grounding, rather than simply increasing context length, is a key bottleneck in document-augmented Text-to-SQL.

\textbf{(3) Agentic workflows perform best overall, but evidence benefits both agentic and standalone models.}
Claude Code achieves the highest Avg accuracy in all three input settings and obtains the best w/Evidence results on Medium and Hard queries.
Meanwhile, Qwen 3.6 Max achieves the best w/Evidence Easy accuracy, showing that concise evidence also benefits standalone LLMs.
Overall, the best Avg accuracy in the full benchmark setting with long-form documents is 15.9\%, highlighting the difficulty of enterprise Text-to-SQL under long-context business knowledge requirements.

\subsection{Practical Expert Ceiling}
\label{subsec:expert-ceiling}

\begin{table}[h]
\centering
\footnotesize
\setlength{\tabcolsep}{3.5pt}
\begin{tabular}{lccc}
\toprule
\textbf{System} 
& \textbf{Base} 
& \textbf{w/Doc} 
& \textbf{w/Evidence} \\
\midrule
Claude Code 
& 6.1 (13) 
& 14.6 (31) 
& 20.3 (43) \\

Expert 
& 33.5 (71) 
& 46.2 (98) 
& 84.0 (178) \\
\midrule
Gap 
& +27.4 
& +31.6 
& +63.7 \\
\bottomrule
\end{tabular}
\caption{
Expert and Claude Code execution accuracy on 212 sampled English examples.
Values in parentheses denote the number of correctly executed examples.
Gap is Expert minus Claude Code.
}
\label{tab:expert-ceiling}
\end{table}

To contextualize model performance against expert-written solutions, we conduct a subset analysis on the 212 sampled English examples.
We view expert performance as a practical reference ceiling rather than an oracle upper bound, since expert-written SQL is still produced under the same enterprise schema, documentation, and business-rule constraints.

Table~\ref{tab:expert-ceiling} shows a substantial gap between Claude Code and expert-written SQL across all input settings.
The gap is especially large under w/Evidence, where expert accuracy increases from 33.5\% to 84.0\%, a gain of 50.5 points over Base, while Claude Code improves from 6.1\% to 20.3\%, a gain of only 14.2 points.
This suggests that localized evidence substantially reduces the knowledge-selection burden for experts, but current agents are still unable to reliably translate localized business rules into executable SQL.
Therefore, the remaining gap reflects not only long-context retrieval difficulty, but also limitations in business-rule grounding, constraint interpretation, and multi-step SQL composition.

\subsection{Domain-wise Analysis}
\label{subsec:domain-analysis}

\begin{figure}[h]
\centering
\includegraphics[width=\linewidth]{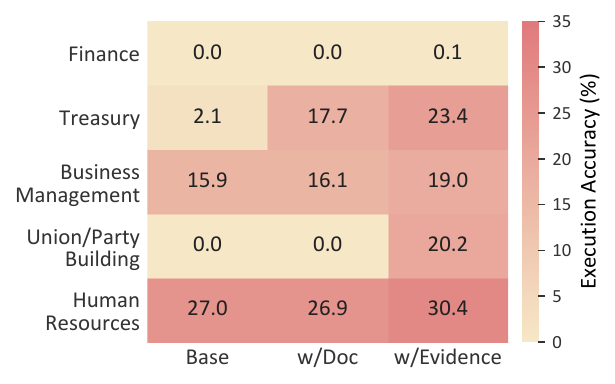}
\caption{
Domain-wise execution accuracy (\%) averaged over all evaluated systems on English examples.
Performance is highly skewed across domains, and concise evidence brings large but uneven gains.
}
\vspace{-10pt}
\label{fig:domain-difficulty}
\end{figure}

Figure~\ref{fig:domain-difficulty} analyzes model performance across enterprise domains.

\textbf{(1) Domain difficulty is highly skewed.}
Performance varies substantially across domains. Under the w/Evidence setting, accuracy ranges from near-zero in Finance to 30.4\% in Human Resources. This indicates that enterprise Text-to-SQL difficulty is not uniformly distributed across domains, but is strongly affected by domain-specific query patterns, business rules, and schema usage.

\textbf{(2) SQL complexity contributes to difficulty, but does not fully explain it.}
Combined with the domain-level SQL-length statistics in Section~\ref{subsec:statistics}, domains with more Easy queries generally achieve higher accuracy, while domains dominated by long and compositional SQL, such as Finance, remain difficult. However, this relationship is not deterministic. For example, Union/Party Building contains no Easy queries, yet its accuracy increases to 20.2\% with w/Evidence. This suggests that structural complexity interacts with how easily the required business knowledge can be localized and grounded.

\textbf{(3) Context helps unevenly across domains.}
Adding long-form documents alone is not consistently effective: it brings a large gain in Treasury, but has little effect in Finance, Business Management, and Union/Party Building, and slightly decreases performance in Human Resources. In contrast, concise evidence provides much larger gains in selected domains, especially Treasury and Union/Party Building. This suggests that models often struggle to locate the relevant rules from long documents, while expert-curated evidence is more useful when the required definitions, scopes, or filtering conditions can be directly mapped to SQL. Nevertheless, the near-zero performance in Finance shows that exposing relevant evidence alone may still be insufficient for domains requiring complex multi-step composition and precise business reasoning.

\subsection{Translation-effect Analysis}
\label{subsec:translation-effect}

\begin{table}[h]
\centering
\scriptsize
\setlength{\tabcolsep}{3pt}
\renewcommand{\arraystretch}{1.08}
\resizebox{\linewidth}{!}{%
\begin{tabular}{@{}l r ccc|ccc|ccc@{}}
\toprule
\multirow{2}{*}{\textbf{Domain}} 
& \multirow{2}{*}{\textbf{\#data}}
& \multicolumn{3}{c|}{\textbf{Base}}
& \multicolumn{3}{c|}{\textbf{w/Doc}}
& \multicolumn{3}{c}{\textbf{w/Evidence}} \\
\cmidrule(lr){3-5}
\cmidrule(lr){6-8}
\cmidrule(lr){9-11}
& 
& \textbf{ZH} & \textbf{EN} & $\boldsymbol{\Delta}$
& \textbf{ZH} & \textbf{EN} & $\boldsymbol{\Delta}$
& \textbf{ZH} & \textbf{EN} & $\boldsymbol{\Delta}$ \\
\midrule
BM 
& 100 
& 16.5 & 15.9 & -0.6
& 17.0 & 16.1 & -0.9
& 18.9 & 19.0 & +0.1 \\
Fin. 
& 351 
& 0.0 & 0.0 & 0.0
& 0.0 & 0.0 & 0.0
& 0.2 & 0.1 & -0.1 \\
\bottomrule
\end{tabular}
}
\caption{
Paired Chinese-English execution accuracy (\%) averaged over all eight evaluated systems.
BM denotes Business Management and Fin. denotes Finance.
$\Delta$ denotes English minus Chinese accuracy.
}
\label{tab:translation-effect}
\end{table}

Although our main quantitative results use English inputs, \benchmark{} also includes aligned Chinese inputs, enabling a controlled analysis of input-language localization effects.
Table~\ref{tab:translation-effect} reports paired Chinese-English execution accuracy on the two domains for which we conduct bilingual evaluation: Business Management and Finance.
For each setting, we average accuracy over all eight evaluated systems and report the English--Chinese difference.
The results show only minor aggregate differences between Chinese and English inputs on these evaluated subsets.
In Business Management, the largest absolute gap is 0.9 percentage points, and the evidence setting yields nearly identical accuracy in the two languages.
Finance remains near zero in both languages across all settings, indicating that its difficulty is dominated by domain and SQL-generation challenges rather than input-language differences.
These findings suggest that translation effects are limited in the evaluated paired subsets, but they should not be interpreted as a comprehensive multilingual evaluation across all domains.

\subsection{Error Analysis}
\label{subsec:error-analysis}

\begin{figure}[h]
\centering
\includegraphics[width=\linewidth]{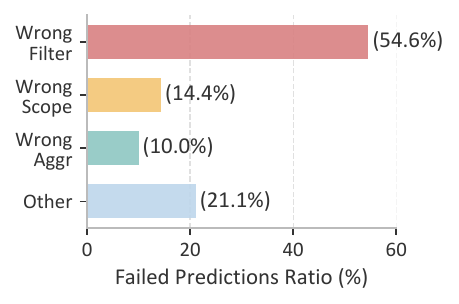}
\caption{
Distribution of primary SQL-level error types among 982 failed predictions from Qwen 3.6 Max under the English input setting.
\textit{Other} aggregates less frequent error types, including syntax, join, calculation, classification, and missing-metric errors.
}
\vspace{-10pt}
\label{fig:error-distribution}
\end{figure}

To better understand model failures, we conduct an error analysis on 982 failed predictions from Qwen 3.6 Max under the English input setting, as summarized in Figure~\ref{fig:error-distribution}. Following prior recommendations for Text-to-SQL error taxonomy design~\citep{liu2025survey}, each failed case is assigned one primary SQL-level error label. Since the analysis is based on a single representative model, it should be interpreted as diagnostic evidence rather than a cross-model conclusion.

\textbf{(1) Constraint capture is the dominant failure mode.}
\textit{WRONG\_FILTER} accounts for 54.6\% of failed cases, making it the most frequent error type. This suggests that many failures are not caused by basic SQL syntax generation, but by incorrect grounding of filtering constraints from the question and domain document. Typical cases include missing required conditions, introducing redundant filters, using incorrect enumeration values, or applying filters to the wrong columns.

\textbf{(2) Models often misidentify the intended data scope.}
\textit{WRONG\_SCOPE} accounts for 14.4\% of failures. These errors occur when the model selects the wrong data range, reporting period, organizational unit, table granularity, or business entity scope. This is particularly problematic in enterprise BI scenarios, where the same metric may depend on specific reporting boundaries or business rules.

\textbf{(3) Aggregation errors indicate remaining weaknesses in SQL composition.}
\textit{WRONG\_AGGREGATION} accounts for 10.0\% of failures. These cases often arise when the model identifies relevant tables or columns but applies an incorrect grouping level, aggregation function, denominator, or intermediate computation. The remaining 21.1\% of errors are grouped as \textit{Other}, indicating that less frequent but diverse failures such as joins, calculations, classifications, and missing metrics still contribute substantially.

Overall, the error analysis shows that current LLM-based Text-to-SQL systems often fail not because they cannot produce valid SQL, but because they struggle to align business constraints, data scope, and metric definitions with the correct SQL operations. These results suggest three promising directions: explicit intermediate representations for complex query planning, stronger grounding to domain-specific business rules, and post-generation validation modules that check schema usage, document constraints, table granularity, and enumeration values.

\section{Related Work}

\textbf{Text-to-SQL Methods.}
Text-to-SQL has evolved from task-specific semantic parsers to LLM-based and agent-centric systems. Early approaches cover core technical directions: seq2seq-based generation (Seq2SQL~\citep{zhong2017seq2sql}, SQLNet~\citep{xu2017sqlnet}), schema-aware linking \& encoding (IRNet~\citep{guo2019towards}, RAT-SQL~\citep{wang2020rat}, Bridging~\citep{lin2020bridging}), sketch-based parsing (SmBoP~\citep{rubin2021smbop}), and constrained decoupled decoding (RESDSQL~\citep{li2023resdsql}). Recent works leverage LLMs, with comprehensive capability evaluation~\citep{gao2023text}, zero-shot optimization (C3~\citep{dong2023c3}, SQLCoder~\citep{defog_sqlcoder_2024}), and decomposed in-context learning with self-correction (DIN-SQL~\citep{pourreza2023din}, DAIL-SQL~\citep{gao2023text}, CHESS~\citep{talaei2024chess}). Multi-agent frameworks further enable iterative schema exploration and error repair (MAC-SQL~\citep{wang2025mac}, RSL-SQL~\citep{cao2024rsl}, SWE-SQL~\citep{li2025swe}). Our work evaluates both standalone LLMs and agent settings, focusing on SQL grounding in long-form enterprise business knowledge rather than only schema matching.

\textbf{Text-to-SQL Benchmarks.}
Text-to-SQL benchmarks have expanded from single-table tasks to cross-domain, conversational, industrial-style evaluation. Early foundational benchmarks include single-table WikiSQL~\citep{zhong2017seq2sql} and cross-domain complex multi-table Spider~\citep{yu2018spider}, with multi-turn conversational extensions SParC~\citep{yu2019sparc} and CoSQL~\citep{yu2019cosql}. Subsequent works broaden evaluation dimensions: multilingual parsing (DuSQL~\citep{wang2020dusql}), external domain knowledge (Spider-DK~\citep{gan2021exploring}), real-world database validity (KaggleDBQA~\citep{lee2021kaggledbqa}), robustness diagnosis (Dr.Spider~\citep{chang2023dr}), and large-scale realistic scenarios (BIRD~\citep{li2023can}). Recent benchmarks further target interactive industrial workflows (BIRD-INTERACT~\citep{huo2026birdinteract}, Spider 2.0~\citep{lei2024spider}). Existing benchmarks overlook enterprise BI's core dependency on private long-form business knowledge, so our \benchmark{} complements them by evaluating business rule-aware SQL generation.

\section{Conclusion}
We introduced \benchmark{}, an enterprise-oriented Text-to-SQL benchmark for evaluating SQL generation grounded in long-form private business knowledge. 
\benchmark{} contains 1,066 aligned Chinese-English semantic examples across five business domains, most of which require external domain knowledge and involve complex gold SQL. 
Experiments on English inputs with strong standalone LLMs and Claude Code show that current systems remain far from reliable in this setting: the best evaluated system reaches 15.9\% when long-form documents are provided. 
Further analysis shows that concise evidence raises the best accuracy in the evidence-ablation setting to 21.4\%, performance varies substantially across domains, and paired Chinese-English subsets show only small aggregate translation effects. 
While we do not claim that \benchmark{} is uniformly harder than prior benchmarks under identical-model evaluation, it exposes a distinct practical gap: faithfully grounding SQL generation in private enterprise knowledge beyond schema matching or short contextual cues. 
We hope \benchmark{} will support future research on more reliable knowledge-grounded Text-to-SQL systems for realistic enterprise scenarios.

\section*{Limitations}
Although \benchmark{} is constructed from real enterprise workloads, it covers only five business intelligence domains from a single enterprise setting. It should therefore not be viewed as representative of all enterprise Text-to-SQL scenarios. Future work can extend \benchmark{} to broader domains, additional database backends, and more realistic workflows, such as multi-turn clarification, permission-aware querying, and latency-sensitive deployment.

Our bilingual setting also has limitations. The English questions, evidence, and documents are translated and validated from the original Chinese materials, so some localization artifacts may remain. As a result, differences between Chinese and English results may reflect both language-specific effects and translation artifacts, and our paired analysis is limited to subsets with available results in both languages. In addition, anonymization may abstract away some organization-specific details that are important in real deployments. Finally, execution accuracy on local evaluation databases does not fully capture robustness, efficiency, access-control behavior, or downstream business utility.

\section*{Ethical Considerations}
\benchmark{} is constructed from enterprise workloads and may involve sensitive business contexts. We therefore apply multi-level anonymization to questions, SQL queries, evidence, documents, and databases before release. Personally identifiable information, organization-specific identifiers, financial institutions, monetary values, and sensitive operational details are removed or replaced with consistent placeholders. Human-resource-related examples are retained only after anonymization and are used for evaluating aggregate Text-to-SQL behavior rather than individual-level profiling. We will release only anonymized data and evaluation scripts, while keeping sensitive gold annotations or proprietary materials private when necessary.

\bibliography{main}

\appendix

\end{document}